\documentclass[letterpaper, 10 pt, conference]{ieeeconf}  

\IEEEoverridecommandlockouts                              

\usepackage{graphics} 

\usepackage{url}
\usepackage{todonotes}

\title{\LARGE \bf
  Insights from Interviews with Teachers and Students on the\\
  Use of a Social Robot 
  in Computer Science Class in Sixth Grade 
}

\author{Ann-Sophie Schenk and Stefan Schiffer and Heqiu Song
\thanks{Chair for Individual and Technology (iTec),
        RWTH Aachen University, 52066 Aachen, Germany
        {\tt\small \{ann-sophie.schenk,stefan.schiffer,heqiu.song\}\linebreak@itec.rwth-aachen.de}}%
}

\begin{document}

\maketitle

\thispagestyle{empty}
\pagestyle{empty}

\begin{abstract}
 In this paper we report on first insights from interviews with teachers and
 students on using social robots in computer science class in sixth grade.
 Our focus is on learning about requirements and potential applications.
 We are particularly interested in getting both perspectives,
 the teachers' and the learners' view on how robots could be
 used and what features they should or should not have. %
 Results show that teachers as well as students are very open to robots in the classroom. However, requirements are partially quite heterogeneous among the groups. This leads to complex design challenges which we discuss at the end of this paper.
\end{abstract}

\section{INTRODUCTION}

Robots have diverse applications across domains such as healthcare, industry, and education.
In the past decade, they have emerged as a promising tool for transforming K-12 education, particularly in STEM (Science, Technology, Engineering, and Mathematics) domains \cite{7428217}.
Recently, computer science education has also gained attention as a key application area for educational robots, driven in part by a global shortage of qualified teachers, which poses challenges for providing individualized instruction.

Against this backdrop, digital technologies have emerged as promising tools to support differentiated instruction.
More specifically, the potential of robots as social learning companions has gained traction in educational research.
Social robots can serve not only as interactive tools but also as intelligent tutoring systems (ITS), autonomous agents capable of adapting to student needs through learner modeling \cite{dillenbourg1992framework, Bull2010, francisco2022intelligent}.
Systems have demonstrated effectiveness in supporting both learners and educators across various domains \cite{vanlehn2011relative, wang2023examining}, including early explorations of Wizard-of-Oz robot tutors in classroom settings \cite{schulz2023towards}. 

In this paper, %
starting from a prototype created in \cite{Schiffer-Liu_HR2024_MoBi-LE},
we explore the use of a socially interactive robot as a didactic mediator in middle school computer science education by interviewing computer science teachers and students from fifth and sixth grade.
To investigate this, we formulated the following research questions: 

\begin{itemize}
    \item How do teachers and students perceive the role of robots in supporting computer science education?
    \item What teaching tasks and learning activities are considered suitable or unsuitable for robotic assistance in the classroom?
    \item What control, interaction, and technical features do users (teachers and students) expect from educational robots?
    \item What concerns or reservations do teachers and students have about integrating robots into the classroom setting?
\end{itemize}

\section{Interviews with Teachers}

In a first event, we tried to get teachers' opinions on using
a social robot to support computer science education in sixth grade.
We attended "Informatiktag NRW", a convention of computer science
teachers of the state of North Rhine-Westphalia in Germany with more than 200 attendees.

\subsection{Procedure}

We had a booth among other exhibitors where teachers passed
by between different sessions and workshops of the event.
We prepared a total of four posters to invite passersby
to engage in a discussion and to contribute their opinion.
The study was approved by the university ethics committee prior to data collection. 

The first poster asked participants to express how \emph{interesting}
and how \emph{useful} they find a social robot in sixth grade
computer science class. %
On a second poster, we asked participants \emph{How could you
imagine using a robot for a computer science class?} and requested
them to put their ideas on sticky notes in one of three columns. %
The first column was used for tasks that should be done
\emph{mainly by the robot}, the second column was meant for tasks
\emph{jointly done by the robot and the teacher}, while the last
column was used for tasks that should be \emph{mainly done by the teacher}. %
A third poster asked for \emph{Which functions should a robot
for a computer science class have or not have?}. %
Finally, the fourth poster asked \emph{How should a robot for a computer science class be designed?}

\begin{figure*}
  \centering
  \includegraphics[width=0.86\linewidth]{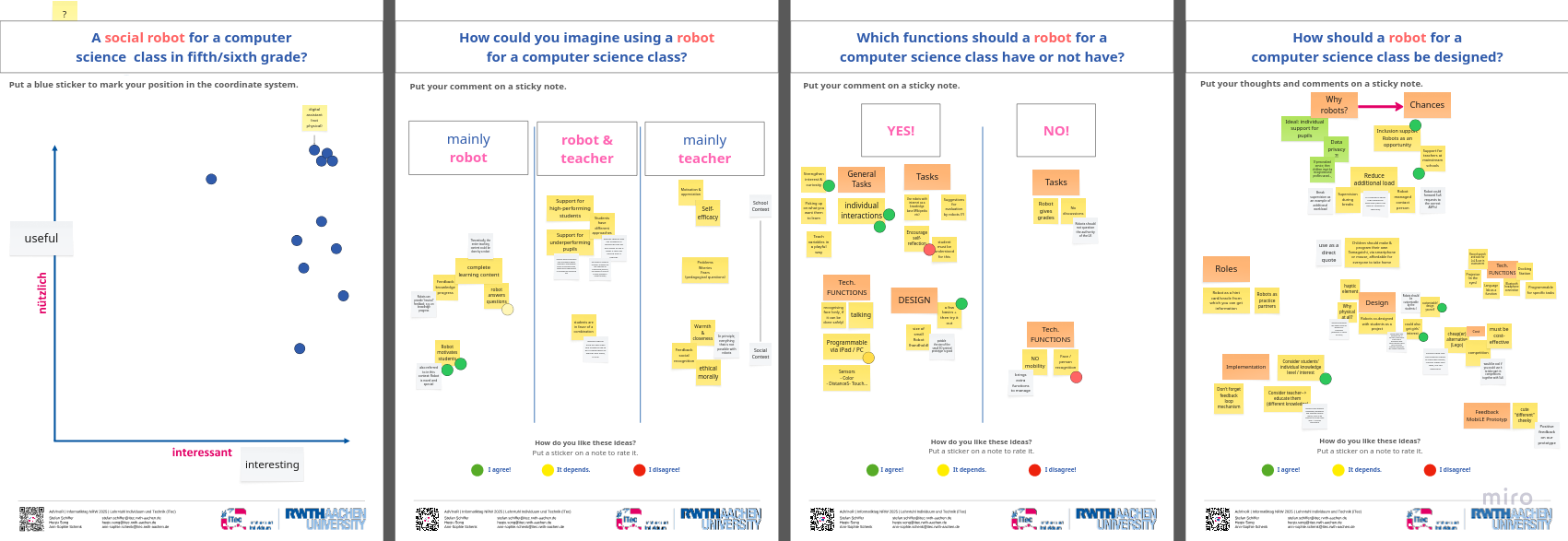}
  \caption{An english translation and digitized version of the four posters
   that we used to collect teachers opinions and feedback.}
   \label{fig:teachers-posters}
\end{figure*}

We had a total of eleven (\textit{N}\,=\,11) teachers as participants all of which answered at least the question of the first poster.
Many of them then engaged in a discussion and either put their own sticky notes on one of the other posters or gave their agreement or disagreement to existing notes.
However, we did not track whether all of them did so.
Figure~\ref{fig:teachers-posters} shows a digitized and translated version of the all the four posters used at the workshop.

\subsection{Results}

We report on results from responses mainly by summarizing the contributions of the individual posters.

\subsubsection{Poster 1}

Overall, participants found the use of a social robot to be interesting, with the majority finding it useful. The absence of anyone in the lower left quadrant (neither useful nor interesting) indicates that participants perceived it as more useful or interesting than not, with many considering both.
Some participants were unsure about the precise meaning of the term "social robot".

\subsubsection{Poster 2}

In terms of what the robot should do on its own, what should be done together with a teacher and what tasks should remain with mainly the teacher participants presented their opinion as follows.

They saw the robot with motivational tasks and would trust the robot with (almost) the complete learning content.
Joint tasks of the robot and the teacher were mainly seen for providing individual personalised support (to account for differences in performance and personality).
One participant mentioned that he would ask his students whether they would prefer a task to be done by the robot, by the teacher, or by both.
As part of the tasks that should rather stay with the teachers many participants mentioned %
motivation and appreciation, %
all sorts of pedagogical questions, and %
any form of emotional support.

It is interesting to note that there was
no agreement among all teachers about who should or should not motivate students %
and that, in general, %
robots should focus more on solving content-related questions
while teachers should focus more on emotional-related questions.
This can somehow be seen as a hard/soft-skill related distinction.

\subsubsection{Poster 3}

With respect to the abilities that a robot should have participants stayed vague,
mentioning that the robot should be able to have individual interactions,
strengthen curiosity and interest, and teach in a playful way.

The few specific functions that were mentioned were that %
the robot should be programmable, %
it should have sensors,
and it should have face recognition to identify individual
students to adapt accordingly, but only if it can be done safely.

\subsubsection{Poster 4}

While the last poster asked for opinions on the design of the robot,
participants used it to give their general feedback beyond the looks.

Among the opportunities that teachers saw two were mentioned several times.
Robots could provide support for teachers, especially with including children with special needs and reducing additional workload.
They are also seen as a cost-effective alternative to support students individually, for example to participate in school competitions or alike.
A concern that what raised often was that a robot should decrease workload and not increase it.

\section{Interview with Students}

In a second event, we used an extra curricular computer science club offered for students from fifth to seventh grade to interview groups of students about their opinions on a socially interactive robot for computer science class.

\subsection{Sample and Procedure}

To explore children's perceptions and expectations regarding the use of social robots in computer science class, five German students (\textit{N}\,=\,5) in 
grades five to seven were interviewed in two groups. 
Although gender was not explicitly assessed, all participants had names conventionally associated with males. Participants were recruited from an existing class in the "Computer Science School Lab" at RWTH Aachen University. 
The parents of the students were informed of the purpose of the study and gave their consent for their children's participation prior to the event.
The study was approved by the university ethics committee prior to data collection. 

\begin{figure}
  \centering
  \includegraphics[width=\linewidth,trim={0 65cm 0 27cm},clip]%
    {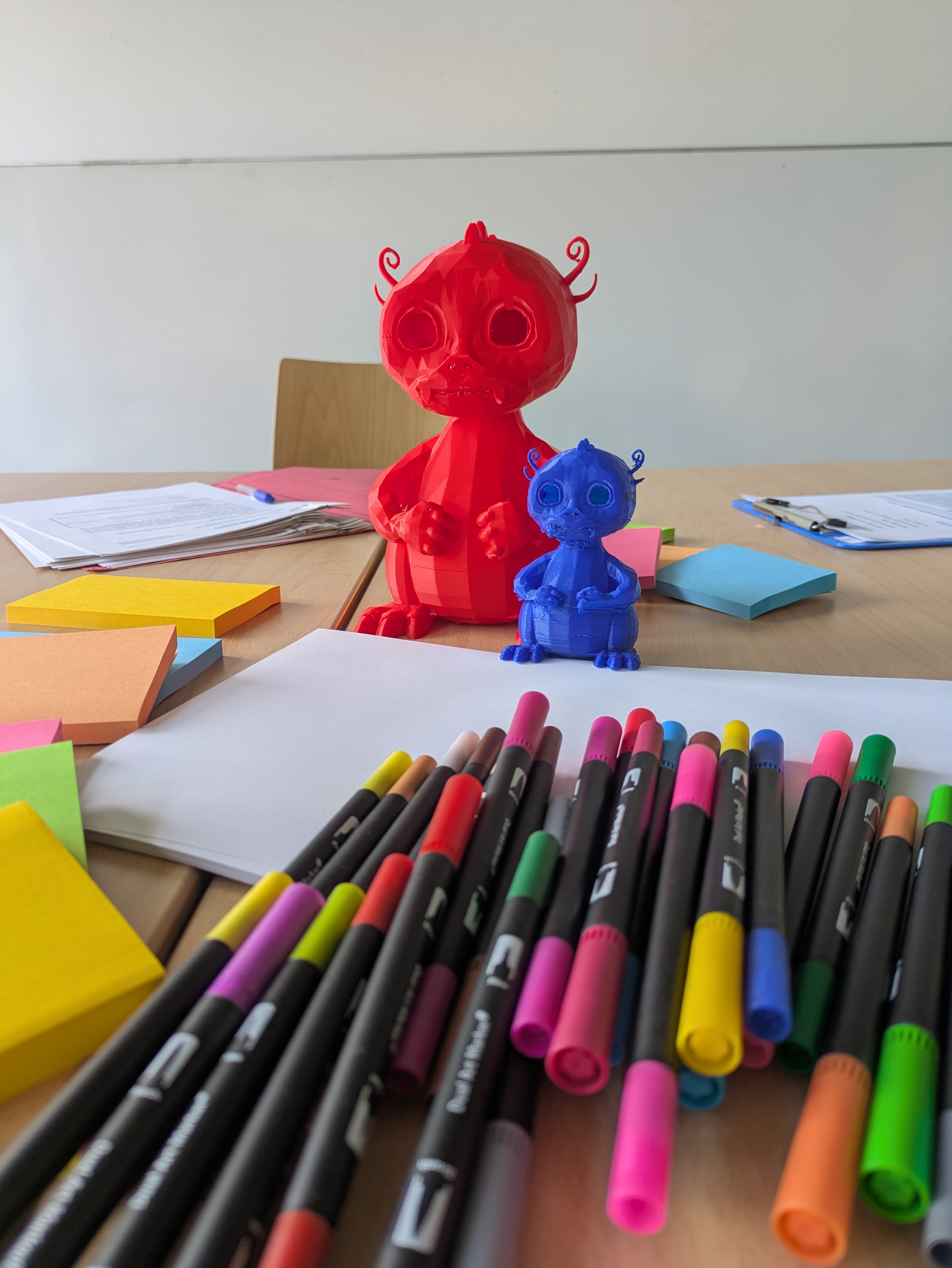}
  \caption{Interview setting showcasing materials used during the group interviews, including colored pens, robot prototypes, sticky notes, and blank sheets of paper.}
  \label{fig:student interview}
\end{figure}

Each group interview lasted approximately 30 minutes. 
We conducted semi-structured interviews, allowing for an optimal degree of standardization while also providing flexibility to adapt questions based on students' responses \cite{SAGE-Encyclopedia-Social-Science-Research-Methods_SemistructuredInterview}.
The guiding questions aimed to elicit students' thoughts on various aspects of robots in computer science education, including their personal definitions of a robot, their opinion on using such a robot in class, desired functionality, and their preferences for customization.
At the end of each session, participants were shown our prototype and asked to share their thoughts.
Figure~\ref{fig:student interview} shows the interview environment with our prototype in two sizes.
To encourage imagination and discussion, they could also draw their own "ideal" robot and describe its appearance and functionalities.
%


\subsection{Results}

Following the group interviews, we transcribed the audio data and conducted reviewed the transcript according to ideas from thematic content analysis to derive insights from the participants' responses.
To establish a common understanding of the topic, we began by asking the children what a robot means to them.
All participants reported having seen one in real life (e.g., vacuum robots, industrial robots in museums, Lego robots).
They defined a robot as "something that can move" (i.e., has arms), "can do something useful", and as "something technical" that does not necessarily "look like something alive".
The students unanimously agreed that robots are diverse and that there is no single "typical" appearance for a robot.
When discussing the potential integration of a robot into their computer science class, they expressed enthusiasm, stating "this would be extremely cool!" and affirming that it "would definitely make computer science class more fun".
However, one student raised concerns about practical issues such as space limitations in their classroom ("there already is not enough space") and the feasibility of using a robot for important tasks ("it depends; a robot should not be used for important tasks because things can go wrong").
Regarding possible tasks for the robot, students suggested its use in assisting with tedious tasks (e.g., finding mistakes) and supporting teachers by answering students' questions.
They also proposed programming challenges where teachers could guide them on how to program their robots to achieve specific goals (e.g., following a straight line).

Interestingly, all students expressed strong opposition to voice output from the robot; they described it as "psycho" and "unnecessary".
Instead, they preferred that the robot understands speech but only speaks when "something is wrong".
In terms of technical features, they desired features such as lights, mobility, cameras, and AI support.
Furthermore, they believed that to effectively answer student questions, the robot should have greater knowledge of computer science than the students themselves. 
The idea of personalizing or customizing the robot was appealing but it was viewed as a secondary feature; they emphasized that functionality was paramount, "most importantly, it should work!".
Basic customization options were also mentioned, including naming the robot and changing light colors ("As a boy who doesn't like pink, I would like to be able to change the light color when it is pink").

The prototype presented was rated as "very cool", with suggestions for an intermediate size between the large and small models.
However, its robustness was questioned; one student commented that "ears are not necessary", identifying them as potential breaking points, while suggesting that it should be "more solid".
The students' drawings of their ideal robots revealed diverse expectations regarding appearance and functionality. Each design illustrated unique use cases that reflected their individual ideas about what robots should look like and the capabilities they should possess.

Overall, students expressed strong enthusiasm for integrating a robot into their computer science class, highlighting its potential to enhance engagement and learning.
However, their wishes, expectations, and proposed use cases were notably diverse, also illustrated by their drawings.
This variability suggests that creating a "one-size-fits-all" solution may be challenging.
Instead, it is crucial to develop a robot that accommodates individual needs and varying levels of knowledge among students to provide optimal support in their learning experiences.

\section{Discussion}

Both, teachers and students generally viewed the use of robots in the classroom as interesting and potentially beneficial. 
Teachers appreciated the potential for small-group interactions using multiple robots, which could enable more personalized learning and better support students with different personalities, preferences, or needs \cite{10.3389/frobt.2021.699524}.
A key expectation was that robots should reduce workload rather than increase it -- favoring simple, stationary designs with limited movement and basic features like LEDs.
Teachers also questioned whether a physical robot was necessary at all, raising concerns about cost-effectiveness.
Their views on the robot's role varied, but many saw robots as suitable for content delivery, while emotional and relational aspects should remain in the domain of human teachers.
Modular, programmable robots were seen as a promising approach to support curriculum flexibility.

Students were also positive about using robots in computer science education.
Their responses revealed diverse expectations and emphasized the importance of personalization.
Interestingly, many were uncomfortable with robots providing verbal feedback -- possibly due to the Uncanny Valley effect \cite{MoriEtAl2012}.
They valued programmable tasks that were engaging and motivating, supporting sustained learning over time.

\subsection{Similarities and Differences between Computer Science Teachers and Students}

Teachers and students shared several views on the potential benefits of classroom robots.
Both groups were generally positive, seeing robots as a way to offer more personalized and individually tailored support.
They agreed on technical preferences such as small size, robustness, and simplicity -- favoring LED indicators over complex displays.
Both appreciated the idea of programmable robots, though with slightly different emphases: teachers focused more on content delivery, while students emphasized interactive tasks that support engagement.
Teachers and students saw value in robots handling repetitive or tedious tasks, such as identifying and correcting errors.

Although there was broad agreement on many aspects, some differing perspectives also became apparent.
For example, teachers preferred robots without mobility, emphasizing practicality and minimal classroom disruption.
In contrast, some students expressed interest in mobile robots, though this feature did not appear to be essential for learning, suggesting that teacher preferences may be more critical for implementation.
Privacy concerns were raised exclusively by teachers, reflecting their heightened awareness of data protection responsibilities.
This is an issue that students may not recognize but which remains important, especially in educational settings.

Verbal interaction was another point of divergence.
Teachers did not object to the idea of a speaking robot and focused more on the quality and type of feedback it should provide.
Students, on the other hand, were clearly opposed to verbal communication, likely due to discomfort with humanoid traits. 
This was supported by their design sketches, which tended to avoid human-like features. 
These findings highlight the need to balance functional interaction with user needs and to remain open to adaptable interaction styles depending on the context.

\subsection{Limitations and Future Work}

This study was exploratory in nature and based on a small, non-diverse, German sample.
As such, findings should be interpreted with caution and may not be generalizable to broader populations.
The next step will be to translate the insights from this study into a functional prototype, followed by iterative user testing.
Involving teachers and students directly, for example, through co-design processes, can be helpful in ensuring that the robot is in line with the real needs and expectations of users.
Future evaluations should specifically examine whether teachers perceive the robot as reducing workload and students as increasing enjoyment and engagement in learning.

\section{Conclusion: Design Insights}

The two studies point to several key considerations for designing educational robots.
Co-design, or citizen science approaches, emerged as particularly valuable: Teachers expressed a strong desire to be involved in the design process and to have the ability to adapt or expand the functions of the robot.
Personalization was important to both groups: Students valued robots that could respond to their preferences and needs, while teachers emphasized the importance of customization to effectively support individual learners.
Design should avoid misleading cues, particularly those that invoke anthropomorphic traits.
Given students' discomfort with humanoid features, nonhumanoid designs are preferable to avoid triggering unrealistic social expectations or discomfort (Uncanny Valley, \cite{MoriEtAl2012}).
Finally, the robot should support playful, game-like interactions, as fun is strongly related to motivation and positive learning attitudes, strengthening its value as a central design principle \cite{TISZA2021100270}.

\addtolength{\textheight}{-12cm}   
\section*{ACKNOWLEDGMENT}

We would like to thank Prof. Nadine Bergner for her support in making it possible to conduct our interviews at Informatiktag NRW and in her Computer Science School Lab.
Also, we gratefully acknowledge the funding received by the exploratory research space of RWTH Aachen University under grant number OPSF834.



\begingroup
  \renewcommand{\baselinestretch}{.92}
  \bibliographystyle{ACM-Reference-Format}
  \bibliography{RO-MAN2025LBR667_AdVIsoR}
\endgroup

%

\end{document}